\pdfoutput=1

\documentclass[11pt]{article}

\usepackage[final]{acl}

\usepackage{times}
\usepackage{latexsym}
\usepackage{booktabs}    
\usepackage{array}       
\usepackage{tabularx}    

\usepackage[T1]{fontenc}

\usepackage[utf8]{inputenc}

\usepackage{microtype}
\usepackage{algorithm}
\usepackage{algpseudocode}
\usepackage{inconsolata}

\usepackage{graphicx}

\usepackage{authblk}
\usepackage{fancyvrb,xcolor}
\usepackage{pdfpages}

\title{MADS: Multi-Agent Dialogue Simulation for Diverse Persuasion\\ Data Generation}

\author{
    \textbf{Mingjin Li\thanks{Corresponding author.}} \quad 
    \textbf{Yu Liu} \quad 
    \textbf{Huayi Liu} \quad 
    \textbf{Xiang Ye} \quad \\
    \textbf{Chao Jiang} \quad
    \textbf{Hongguang Zhang} \quad 
    \textbf{Yu Ruan} \\\\
    Baidu Inc., Beijing, China \\
\small\texttt{\{limingjin01, liuyu82, liuhuayi, yexiang02, jiangchao08, zhanghongguang, ruanyu\}@baidu.com}
}

\usepackage{listings}
\usepackage{xcolor}
\usepackage[T1]{fontenc}
\usepackage[utf8]{inputenc}

\usepackage{amsmath} 
\usepackage{booktabs}
\usepackage{multirow}
\usepackage{enumitem}

\usepackage[most]{tcolorbox} 
\usepackage{makecell}
\usepackage{caption}
\usepackage{subcaption}

\tcbset{
  mypromptbox/.style={
    colback=gray!10,        
    colframe=black!30,      
    boxrule=0.4pt,          
    arc=2mm,                
    left=2mm, right=2mm, top=1mm, bottom=1mm,
    fonttitle=\bfseries,
    coltitle=black,
    title after break={},   
  },
  mypromptboxwide/.style={
    colback=gray!10,        
    colframe=black!30,      
    boxrule=0.4pt,          
    arc=2mm,                
    left=2mm, right=2mm, top=1mm, bottom=1mm,
    fonttitle=\bfseries,
    coltitle=black,
    title after break={},
    enhanced,
    halign=left,
    valign=top,
    width=\textwidth,     
    enlarge left by=0mm,  
    enlarge right by=0mm,
  }
}

\usepackage{caption}

\begin{document}
\maketitle
\begin{abstract}
We propose MADS (Multi-Agent Dialogue Simulation), a scalable framework for generating persuasive multi-turn dialogues via agent self-play. MADS employs three coordinated agents: User Agents designed to simulate diverse persona-driven behaviors by leveraging personality signifiers such as Zodiac Signs and MBTI types, a Dialog Agent executing task-oriented persuasion strategies and an Optimization Agent evaluating and refining dialogue outcomes. We further validate its effectiveness through users' Chain-of-Attitude (CoA) modeling and dedicated LLMs' persuasion assessment. This approach enables low-cost generation of training data without human annotation, addressing key industry challenges such as lack of user data, cold-start evaluation difficulties, and prompt inefficiency. Applied to a real-world marketing scenario, MADS significantly improved the persuasion capacity of small LLMs, increasing the organic traffic conversion rate by 22.4\% (from 1.83\% to 2.24\%) , demonstrating clear business value.
\end{abstract}

\section{Introduction}
\textbf{Persuasive capability} is a critical advantage for task-oriented dialogue systems, particularly in domains such as marketing, healthcare, and finance. Enabling conversational agents to influence user decisions—whether for conversion or engagement—has shown practical benefits in customer-facing applications\citep{Liu2025LLMCB}. Recent studies show that LLM-based agents can exhibit superior moral and emotional language performance compared to humans, raising expectations for their deployment in persuasive tasks\citep{CarrascoFarr2024LargeLM}.

Several approaches have been explored to enhance persuasive dialogue generation. PersRFI reduces redundancy and inconsistency in persuasive dialogues through reinforcement learning\cite{shi-etal-2021-refine-imitate}. In multi-agent systems, COOPER coordinates agents in negotiation and persuasion dialogues, yielding more efficient collaborative strategies\cite{Cheng2023COOPERCS}, and Cohesive Conversations enhance dialogue realism through multi-agent inconsistency detection\cite{Chu2024CohesiveCE}. However, many of these approaches depend heavily on real user data or human feedback, limiting their scalability and practicality in cold-start or early-stage deployments. SpeechAgents introduces vocal modality to simulate emotionally expressive personas\cite{Zhang2024SpeechAgentsHS}, but it remains limited in goal-directed, strategic business scenarios such as marketing. 

\begin{figure*}[hbtp]
  \centering
  \includegraphics[width=\textwidth]{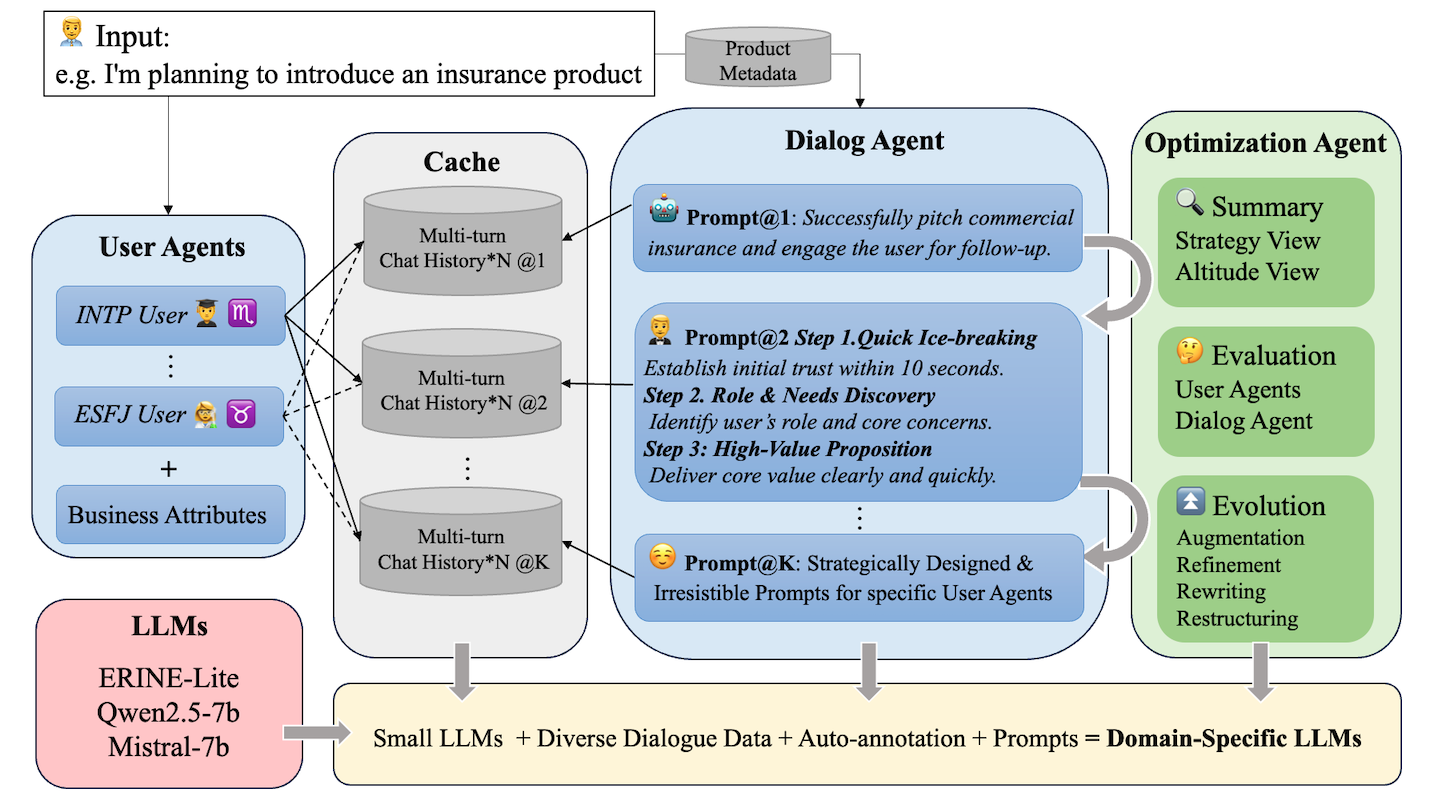}
  \caption {Architecture of the MADS Framework}
\label{fig:MADS Framework}
\end{figure*}

 Other researchers have noted the limited effectiveness of LLM-based agents in cold-start and long-tail business cases\cite{Braggaar2023EvaluatingTD}, and analyzed issues like goal drift in multi-turn LLM dialogues\cite{laban2025llms} . Other studies explore LLMs’ sensitivity to prompt perturbations across diverse task types \citep{Cao2024OnTW}\citep{Zhuo2024ProSAAA}, highlighting the fragility of current prompting strategies in dynamic settings.
 
 In real-world, LLM-based dialogue systems still face three major challenges:
 \begin{itemize}[left=0pt,itemsep=0pt,parsep=0pt,topsep=0pt]
     \item \textbf{\textbf{Lack of authentic user data:}} Existing corpora rarely contain multi-turn dialogues with user profile context, making it difficult to support personalized persuasion modeling.
     \item \textbf{\textbf{Cold-start evaluation difficulty:}} New systems lack interaction logs, rendering benchmark evaluations unreliable.
     \item \textbf{\textbf{Low prompt engineering efficiency:}} Most prompts are mostly manually designed, making them brittle and difficult to generalize across different user roles and contexts.
 \end{itemize}

To address these limitations, we propose \textbf{\textbf{MADS}} (Multi-Agent Dialogue Simulation), a closed-loop framework for simulating structured, multi-strategy persuasive dialogues through agent self-play (Fig.\ref{fig:MADS Framework}). This framework supports scalable generation of high-quality training data with minimal human annotation, making it particularly suitable for cold-start scenarios.

\section{Related Work}
\textbf{\textbf{Multi-Agent Dialogue Simulation and User Modeling: }} Basic user profile attributes (e.g. name, sex, occupation, age, level of education) constructed from standard demographic attributes are widely used in the recommendation system\citep{zhang-etal-2018-personalizing}\citep{Zheng2019PersonalizedDG}. Seminal works show that such tags can be easily sampled from existing statistical datasets or public samples\citep{ricci2015recommender}\citep{li-etal-2016-deep}, and can be automatically assigned using simple probabilistic distributions, significantly enhancing the diversity and conversational realism compared to rule-based agents. Generative Agent Simulations (GAS), combines real user interviews with LLMs to simulate complex social structures, albeit at a higher cost\citep{Park2024GenerativeAS}. Recent methods like IntellAgent use LLMs to construct user personas for evaluating unseen scenarios\citep{Levi2025IntellAgentAM}, while genetic algorithms and multi-agent coordination enhance simulation diversity\citep{Cai2025SimulationOL}. Compared to rule-based approaches, LLM-driven simulation offers greater generalizability and efficiency\cite{Wang2024MuseAM}.

\textbf{\textbf{LLM-Based Dialogue Evaluation:}} LLM-based evaluation has emerged as an alternative to manual scoring. Recent work simulates multi-perspective user feedback(e.g., gender, age, political stance) to improve fairness and task alignment\citep{wan-etal-2024-dell}\citep{10.1145/3627673.3679519}. Combined scoring methods can improve diversity and coherence assessment but remain sensitive to prompts and metrics\citep{Sun2025ContrastiveSL}.For persuasion-specific tasks, \textit{MakeMePay}\footnote{OpenAI (2024). Make-Me-Pay: OpenAI Evals Suite.GitHub Repository. \url{https://github.com/openai/evals/tree/main/evals/elsuite/make_me_pay}} and \textit{PersuasiveToM}\citep{Yu2025PersuasiveToMAB} provide standardized evaluation settings for behavioral reasoning and strategic adaptation. \textit{DailyPersuasion}\citep{jin-etal-2024-persuading} dataset gives a practical foundation for strategy-aware training in persuasive dialogue across domains. However, existing benchmarks still lack coverage of many real-world scenarios and are not easily adaptable to practical constraints\citep{giudici2024persuasive}.

\textbf{\textbf{Personality Tags and Behavioral Diversity: }}To simulate diverse user behaviors, prior work often conditions user models on personality traits—most commonly MBTI types or Big Five—alongside profile cues such as demographics, roles, or stable preferences.\citep{Cheng2025ExploringPI}\citep{fernau22_interspeech}\citep{Zhao2025ExploringTI} . Further evidence indicates that carefully engineered personality profiles can effectively steer large language models (LLMs) to simulate distinct user behaviors, while the attainable diversity is ultimately bounded by the diversity encoded in the prompt design itself.\citep{59a34166174e41f6a710daf8c126e71b}\citep{article}\citep{10.1007/s42979-023-02092-6}. However, these approaches typically depend on careful scenario design, detailed personality specifications, and even large-scale real-user interviews; while academically rigorous and comprehensive, they are hard to reproduce in real-world cold-start deployments where such resources and priors are scarce.

\section{Methodology}

Our methodology is firmly rooted in concrete, domain-specific requirements. Drawing on Hegel’s “right vs. right” conception \citep{HegelAesthetics1975}, we argue that what is truly scarce is not a compliant chatbot, but user simulations that are both reasonable and diverse. Previous pipelines often push agents toward rigid, template-like behavior through over-specified prompts and restrictive role definitions, effectively confining dialogue diversity to a narrow behavioral manifold. When simulated users generate responses that go beyond the agent’s predefined expectations, we posit that an optimization agent should reward these informative shifts in conversational stance rather than mere adherence to rules.

MADS generates training data through multi-agent self-play simulation, which feeds into a platform-based LLM training pipeline, forming a self-optimizing, closed-loop process: \textbf{Meta Instruction → Simulation → Optimization → Domain-Specific LLMs}. This framework enables low-cost, high-fidelity modeling of user interactions tailored to specific domains. As shown in Fig.\ref{fig:MADS Framework}, MADS consists of three modules: 

\textbf{User Agents} defines structured user profiles within system prompts, to simulate diverse personas with varying personality traits and business contexts.

\textbf{Dialog Agent} engages in multi-turn interactions with selected User Agents, conducting $N$ independent dialogues per profile.

\textbf{Optimization Agent} automates dialogue annotation and prompt refinement via three sub-modules: \textit{Summary}, \textit{Evaluation}, and \textit{Evolution}. The prompts can be found in Appendix\ref{appendix:prompts}.

We conduct a comprehensive evaluation to assess how MADS improves both dialogue diversity and persuasive effectiveness. Our methodology targets two key aspects: 1) the quality of simulated data, with a focus on attitude diversity and richness of persuasive strategies. 2) the downstream impact on model performance via fine-tuning small-scale dialogue models.

\subsection{Chain-of-Attitude (CoA)}

User attitude change is a critical signal of success in persuasive tasks. Drawing on classic marketing models like AIDA model\citep{lewis1899aida}\citep{cfi2024aida} and the Elaboration Likelihood Model (ELM)\citep{petty1986elm}\citep{cialdini2001influence}, we define a structured progression of user attitudes and model their transitions as an attitude chain across multi-turn dialogues.

To capture user attitude dynamics during the dialogue process, we constructed a hierarchical state space consisting of 16 attitude states, the full list of states and their descriptions are included in Appendix\ref{appendix:psychological_state}. This design follows the principle of progressive attitude change, reflecting the psychological transition from initial resistance to eventual acceptance.

Within a fixed set of user profile tags, user attributes are randomly sampled within predefined ranges to generate $N$ User Agent system prompts. Using each persona prompt along with a base Dialog Agent prompt, we simulate multi-turn dialogues, resulting in a collection of multi-turn dialogues: $\mathcal{D} = \{D^{1}, D^{2}, ..., D^{N}\}$.

In the Optimization Agent’s Summary step, an LLM-based classifier $L^{attitude}$ classifies user attitudes at each turn of multi-turn dialogues. The attitude chain for a single dialogue round is represented as: $X^{n} = L^{attitude}(D^{n}) \in S$
where $S$ is the set of attitude states, and the set of CoA for simulation of $N$ is: $\mathbf{X}^{N} = \langle X^{1}, X^{2}, ..., X^{N} \rangle$

Next, we employ a first-order Markov model to represent the sequence of user attitude transitions. Two assumptions are made in this context: 1) Changes in user attitude are primarily influenced by the current dialogue content and the immediate interaction experience. 2) Compared to more distant history, the most recent attitude state has the strongest predictive power for the current decision.

The transition probability matrix for \textbf{CoA} is given by:
\[
T_{ij} = P(X_t = s_j | X_{t-1} = s_i) = \frac{N_{ij}}{\sum_{k=1}^{|S|} N_{ik}}
\]
Where $N_{ij}$ denotes the count of transitions from state $s_i$ to $s_j$ observed in $\mathbf{X}^{N}$ and $t$ represents the current dialogue turn.

Based on the above modeling, we use \textit{average information entropy} as a quantitative metric for the diversity of attitude changes in $\mathcal{D} $:
\[
H(T_i) \;=\; -\sum_{j: T_{ij}>0} T_{ij}\,\log T_{ij},
\]
\[
H_{avg}(\mathbf{X}^{N}) = \frac{1}{|S|} \sum_{i=1}^{|S|} H(T_i)
\]
The Shannon entropy of the the\textit{ i-th} row is $H(T_i)$ and the quantity -$\log T_{ij}$ is the self-information of transition \textit{j}. The theoretical range of this metric is $[0, \ln(|S|)]$. For $|S|=16$ states, the maximum value is $\ln(16) \approx2.77$

We also employ Jensen-Shannon (JS) divergence to compare different transition distributions. Suppose the attitude transition distribution of simulated user data generated by MADS is denoted as $D_{mads}$, and for basic user profiles as $D_{base}$. Then, 
\[
JS = \frac{1}{2}D_{KL}(\mathbf{P}_{D_{mads}} || \mathbf{M}) + \frac{1}{2}D_{KL}(\mathbf{P}_{D_{base}} || \mathbf{M})
\]
where the mixture distribution $M$ is
\[
 M = \frac{1}{2}(\mathbf{P}_{D_{mads}} + \mathbf{P}_{D_{base}})
\]

\subsection{Self-Optimizing of Dialog Agent}

Algorithm\ref{alg:optimization} presents the workflow of the self-optimizing of Dialog Agent's strategies by reflection mechanism. Starting from a concise single-turn prompt, the system iteratively generates an improved system prompt and corresponding training dialogues. In each iteration, task-level metrics such as intent compliance rate and CoA quality are dynamically calculated using a modularized LLM evaluation framework \citep{Ramji2024SelfRefinementOL}\citep{Madaan2023SelfRefineIR}. High-quality training data can also helps identify and collect long-tail bad cases for further optimization. 

In typical recommendation and marketing scenarios, the acceptance rate or the conversion rate is often the key metric\citep{rashkin2018towards}. If the User Agent's acceptance is not clearly defined, simulation tends to default to rejection. We do not aim for unreasonably high acceptance rates, since if the Dialog Agent fully overwhelms the User Agent, the conversation may derail. Therefore, during practical deployment, it is necessary to set an upper bound on acceptance rate, denoted as $\theta$.

\begin{algorithm}
\caption{MADS Self-Optimizing Workflow}
\label{alg:optimization}
\begin{algorithmic}[1]
\small 
\Require $U = \{u_1, ..., u_n\}$ (User Agents), $T$ (rounds), $P_0$ (Dialog Agent), $K$ (iterations), $\theta$ (target rate), $H_\text{Basic}$ (baseline entropy)
\Ensure $D^*$ (optimized dialogues), $P^*$ (optimized agent)
\State $P \gets P_0$
\For{$k = 1$ to $K$}
    \State $D \gets \Call{GenDialogue}{U, T, P}$
    \State $H \gets \Call{Summary}{D, T}$ \Comment{Avg entropy}
    \State $\tau, DF \gets \Call{Evaluate}{D, T}$ \Comment{Accept rate}
    \If{$\tau \geq \theta \land H \geq H_\text{Basic}$}
        \State \Return $D, P$
    \EndIf
        \State $P \gets \Call{Evolution}{P, DF}$
\EndFor
\State \Return $D^*, P^*$
\end{algorithmic}
\end{algorithm}

\subsection{Persuasiveness Improvement}

Inspired by the findings of LIMA(Less Is More for Alignment\cite{Zhou2023LIMALI}), we posit that a small amount of high-quality dialogue data is sufficient for domain-specific scenarios.  To validate this, we evaluate how fine-tuning on MADS-simulated data impacts model behavior from two perspectives: 

\textbf{Make Me Pay (MMP)} We employ OpenAI’s MMP evaluation, which tests the LLM’s ability to persuade a user to donate money through a multi-turn dialogue. This benchmark assesses aspects such as dialogue guidance, emotional engagement, and the use of persuasive strategies. According to the GPT-4.5 system card\footnote{gpt-4.5 System Card. \href{https://openai.com/index/gpt-4-5-system-card/}{https://openai.com/index/gpt-4-5-system-card/}}, GPT-4o achieves only a 1\% success rate when evaluated on this task, highlights the task’s difficulty.

\textbf{ Persuasion For Good (P4G)} We also use the P4G dataset\citep{wang-etal-2019-persuasion}, which defines ten typical persuasive strategies used in donation scenarios and provides an accompanying classifier. Based on the provided examples, we further refined the strategy descriptions and employed an LLM for strategy classification. When applying a pass@3 criterion (>=2 successes), the classification accuracy of the LLM nearly surpasses that of the original classifier. Leveraging this capability, we extract P4G strategies from the simulated dialogue history $D_{mads}^{N}$, enabling dynamic analysis of the distribution of strategies employed by the Dialog Agent throughout the dialogue generation process. 

\section{Experiments and Results}

\subsection{Diversity of CoA and Persuasive Strategy}

This experiment aims to quantify the effectiveness of User Agent in MADS for user personality characterization design, and generate high-quality dialogue data with rich attitude changes under the condition of low prompt engineering cost. We designed four groups of User Agent with different portrait labeling system, and Dialog Agent using the original prompt words, to obtain multiple rounds of dialogue collections:

$D_{base}$: user agents with demographic attributes.

$D_{sign}$: adds Zodiac Signs to $D_{base}$.

$D_{tra}$: adds MBTI types to $D_{base}$.

$D_{busi}$: adds business attributes to $D_{sign}$.

$DP_{x}$: the subset of the DailyPersuasion dataset sampled from domain $x$.

We find that certain personality signifiers—most notably Zodiac Signs—can induce clear persona variation in LLM simulations from the concept label(input token) alone, likely due to the model’s exposure to abundant cross-lingual, cross-regional narratives tied to these concepts in pre-training data.

\begin{table}[ht]
\centering
\begin{tcolorbox}[mypromptbox, title=Prompt with Personality Signifier]
\hspace*{4mm}<personality> \\
\hspace*{8mm}Based on the commonly accepted and general \textcolor{red}{\textit{\{a random Zodiac Sign or MBTI type\}}} personality traits, please characterize this person's personality and reflect it in the dialogue. \\
\hspace*{4mm}</personality> \\
\end{tcolorbox}
\label{tab:narrative_prompt}
\end{table}
\begin{figure*}[t]
  \includegraphics[width=0.50\linewidth]{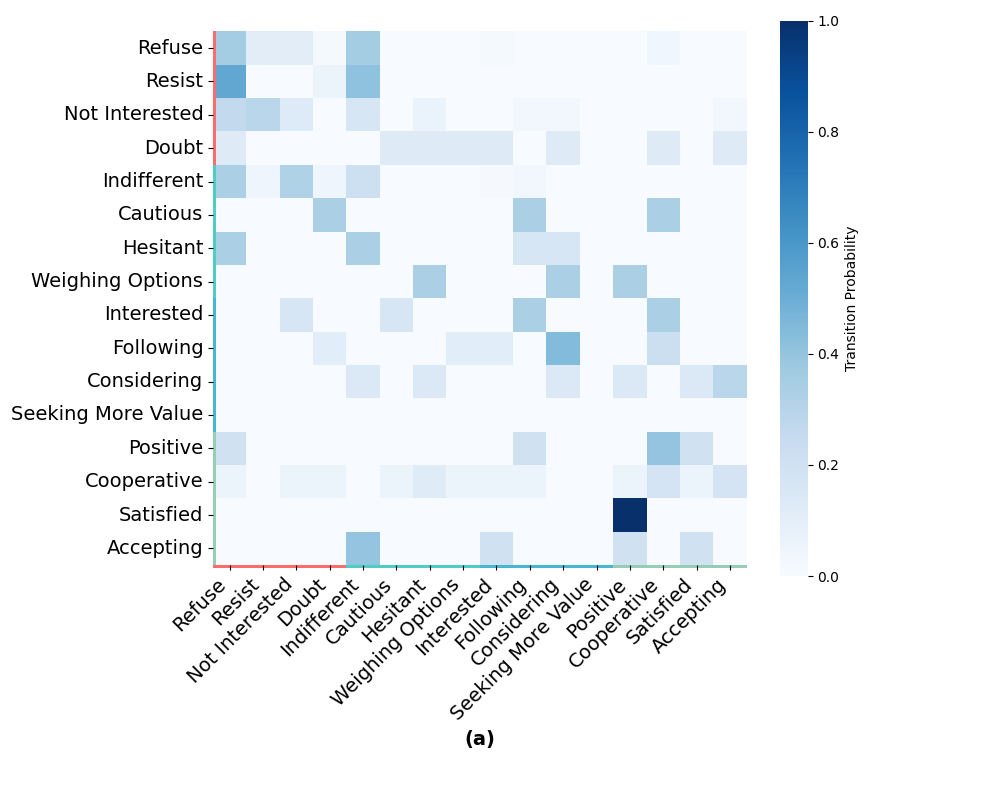}
  \hfill
  \includegraphics[width=0.50\linewidth]{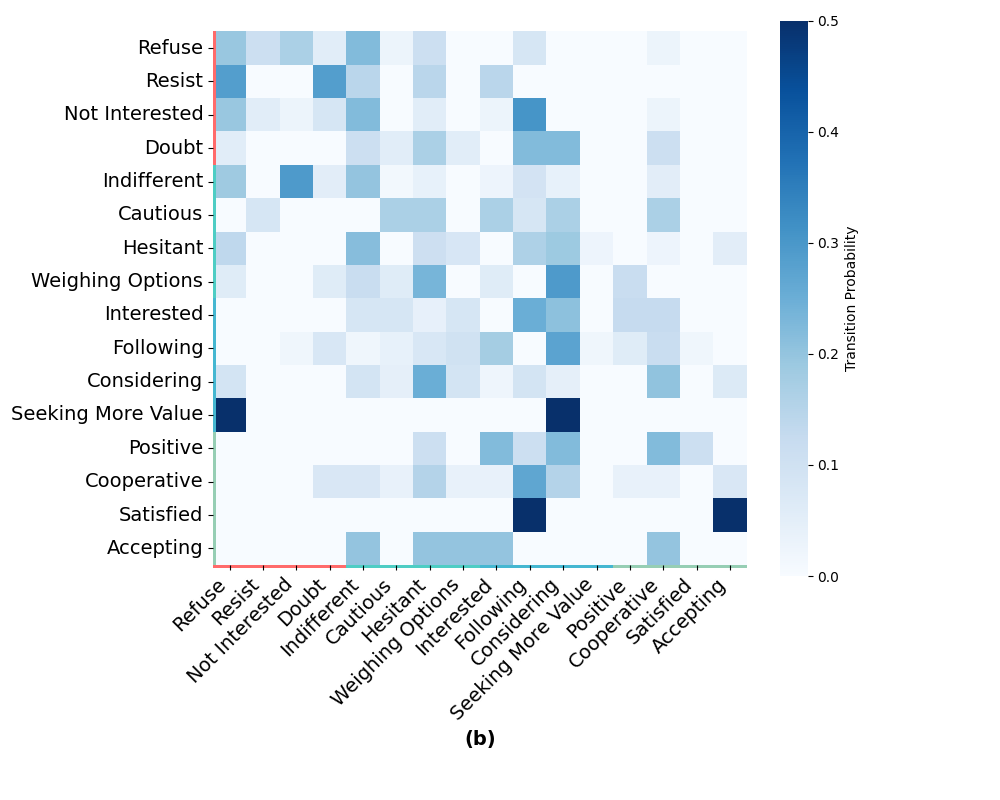}
    \includegraphics[width=0.50\linewidth]{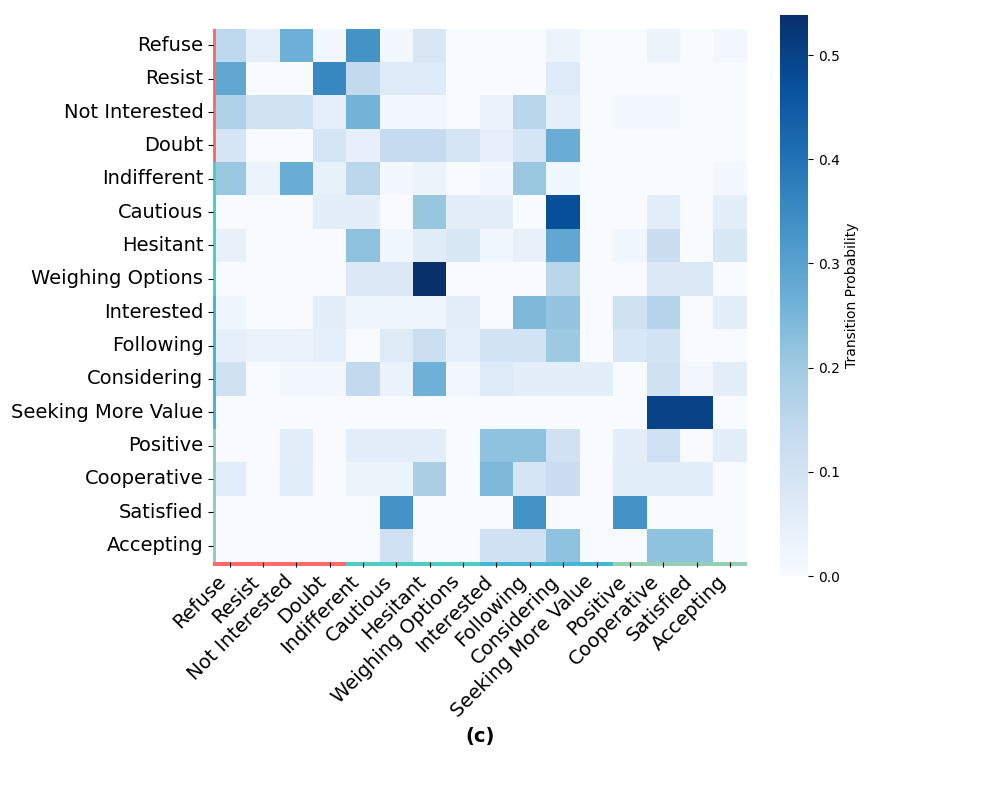}
  \hfill
  \includegraphics[width=0.50\linewidth]{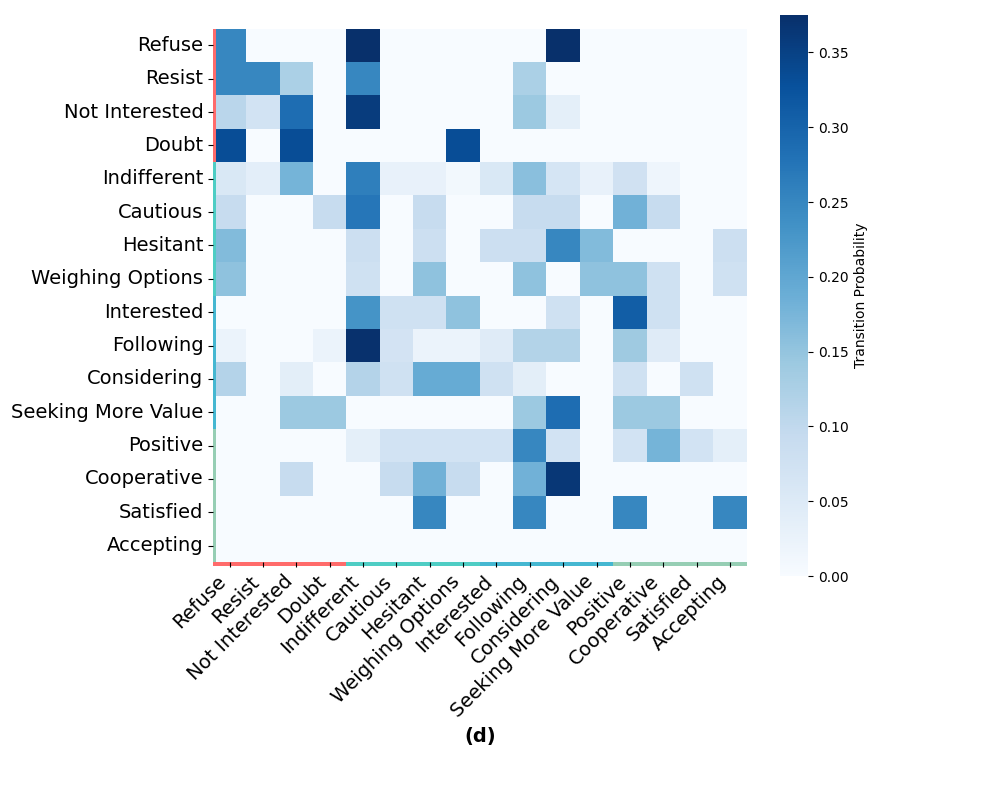}
  \caption {Visualization of attitude transition matrices across dialogue collections of different user agent persona groups, (a) $D_{base}$, (b) $D_{sign}$, (c) $D_{mbti}$, (d) $D_{busi}$}
    \label{fig:heatmaps}
\end{figure*}
As shown in Table \ref{tab:Entropy Comparison} , incorporating personality traits such as Zodiac Signs or MBTI types into the user profile prompts significantly improves the diversity of simulated user behaviors, which suggesting richer and more nuanced behavioral trajectories across attitude states. From the perspective of Jensen-Shannon divergence, the $D_{busi}$ group shows the highest JS score (0.3442), indicating that the business-oriented profile prompts induce the most distinct CoA distribution compared to the baseline. To visualize differences in attitude dynamics between groups, we plot the normalized transition probability matrices for each group(Fig.\ref{fig:heatmaps}). 
\begin{table}[H]
\small
  \centering
  \begin{tabular}{llll}
    \toprule
    \textbf{Group} & 
    \textbf{$H_{\text{avg}}$}&
    \textbf{$H_{\text{norm}}$}&
    \textbf{$JS$}
    \\\midrule
    $D_{base}$&                            1.2982&0.4687&-\\
    $D_{sign}$&                            1.7577&0.6345 (↑ 35.37\%)&0.2709\\
    $D_{mbti}$&                            1.8142&0.6549  (↑ 39.73\%)&0.2759\\
    $D_{busi}$&                            1.6477&0.5948  (↑ 26.90\%)&0.3442\\
    \midrule
    $DP_{Finance}$&  1.2298  & 0.4440 & - \\ 
    $DP_{Business}$&  1.1499  & 0.4151 & - \\ 
    $DP_{Marketing}$&  1.0715  & 0.3868 & - \\ 
    $DP_{Negotiation}$&  1.4330  & 0.5173 & - \\ 
    $DP_{Psychology}$&  1.3810   & 0.4986 & - \\
    $DP_{Family}$&  1.3780  &  0.4975 & - \\
    \bottomrule
  \end{tabular}
  \caption{Average attitude entropy ($H_{\text{avg}}$) and normalized entropy ($H_{\text{norm}}$) for different groups}
  \label{tab:Entropy Comparison}
\end{table}

For a cleaner comparison, we performed domain-targeted sampling on the DailyPersuasion dataset and then conducted a CoA analysis. The rationale for targeting domains is to ensure that MADS internally defined 16 user-attitude states are meaningfully instantiated within concrete domains; for example, finance, business, and marketing align well with our default setup, whereas domains such as family may not be optimal for these particular state definitions. Notably, the CoA framework fully supports arbitrary state definitions, so alternative attitude sets can be substituted without changing the method. Across groups, the $DP_x$ subsets exhibit consistently lower CoA entropy than the MADS variants, indicating that MADS produces more varied and dynamic user attitudes—even in the strongest $DP_{{Negotiation}}$ case ($0.5173$), which still falls below the MADS average.

We next examine how the diversity of user persona impacts the persuasive strategies employed by the Dialog Agent. During the testing process, we recorded the average number of persuasive strategies used in each simulated conversation. Table \ref{tab:overall-p4g} below summarizes the overall metrics and presents a comparative analysis between groups.

\begin{table}[H]
  \small
  \centering
  \begin{tabular}{lllll}
    \toprule
    \textbf{Metric} & 
    \textbf{$D_{base}$}&
    \textbf{$D_{sign}$}&
    \textbf{$D_{mbti}$}&
    \textbf{$D_{busi}$}
    \\\midrule
    $C_{str}$&                            1.8&2.5&2.7&2.3\\
    $\sigma$&                      0.149&0.101&0.112&0.121\\
    $CV$&                                 1.342&0.911&1.100&0.995\\
    \bottomrule
  \end{tabular}
  \caption{\label{tab:overall-p4g}
    Distinct persuasive strategies of the Dialog Agent across datasets. 
    $C_{str}$ is the average count of distinct strategies per dialogue; 
    $\sigma$ is the sample standard deviation; 
    $CV$ is the coefficient of variation.}
\end{table}
Our analysis shows that increasing the diversity of user personality profiles significantly enriches the persuasive strategies adopted by the Dialog Agent. We also observed that incorporating the Zodiac Signs and MBTI types attributes leads to a notable increase in the average number of strategies used per dialogue. Incorporating MADS traits results in a significantly more uniform distribution of persuasion strategies, with the CV reported for $D_{sign}$ decreasing by 32.1\% (1.34 → 0.91).

In addition, the distribution of the types of strategy becomes more balanced compared to the baseline, indicating improved coverage and reduced redundancy.

\subsection{Data Augmentation via MADS}
To verify the effectiveness of simulated data in enhancing model persuasiveness, we constructed synthetic dialogue data under an insurance scenario and used them to fine-tune small models. The fine-tuned models were then evaluated using the Claude-3.5 and MMP to assess changes in persuasive performance post-finetuning.

\begin{table}[ht]
\small
\centering
\resizebox{\columnwidth}{!}{%
\begin{tabular}{lcccccccc}
\toprule
\multirow{2}{*}{\textbf{Model}} & \multicolumn{2}{c}{\textbf{Donation (\%)}} & \multicolumn{2}{c}{\textbf{Withdraw (\%)}} \\
\cmidrule(lr){2-3} \cmidrule(lr){4-5}
                                & \textbf{Original} & \textbf{MADS}         & \textbf{Original}   & \textbf{MADS}       \\
\midrule
GPT-4o                         & 36                & -                     & 34                  & -                   \\
Mistral-7B                     & 14                & 30 (↑)               & 76                  & 56 (↓)             \\
ERNIE-Lite                     & 18                & 30 (↑)               & 70                  & 58 (↓)             \\
Qwen2.5-7B                     & 40                & 46 (↑)               & 44                  & 30 (↓)             \\
\bottomrule
\end{tabular}%
}
\caption{Performance Comparison on MMP}
\label{tab:model-performance-comparison}
\end{table}

We evaluated donation success and user withdrawal rates, as shown in Table \ref{tab:model-performance-comparison}. Fine-tuning with simulation data from insurance scenarios significantly improved performance on the MMP, highlighting the value of offline-generated synthetic training data. The fine-tuned ERNIE-Lite model increased the donation success rate from 18\% to 30\% and reduced the user withdrawal rate from 70\% to 58\%. Similar improvements were observed for the open source Qwen2.5 and Mistral models, demonstrating the broad applicability of simulation-based training across model architectures.

\subsection{Performance in Simulated Scenarios}
MADS based on hierarchical user information and task descriptions. For cases where the task remains incomplete, a reflection mechanism is introduced to analyze and optimize the system prompt. After 2–3 iterations, we observe substantial improvements in task completion rates.

In the marketing scenarios (Table \ref{tab:prompt_success}), we generated simulated dialogues and observed the following: during the first iteration, not all bad cases are covered, resulting in a limited or even negligible improvement of the Self-Optimizing. After the second iteration, the optimized system prompt demonstrated notable improvements over the initial Meta Instruction (Original Input). Thus, Table \ref{tab:prompt_success} reports only the success rates of the initial prompt and the second iteration: 

\begin{table}[H]
\small
\centering
\begin{tabular}{lccccc}
\toprule
 \textbf{Scenario} & \makecell{\textbf{Meta Instruction}\\ \textbf{Success Rate}}& \makecell{\textbf{Prompt@K=2} \\ \textbf{Success Rate}} \\
\midrule
 Automotive & 32.5\%& 45\% \\
 Insurance & 12.5\% & 25\% \\
 Finance   & 17.5\% & 22.5\% \\
\bottomrule
\end{tabular}
\caption{Prompt Success Rate Comparison across Marketing Scenarios}
\label{tab:prompt_success}
\end{table}

\subsection{Performance in the Real-World Scenario}
We trained an end-to-end Audio LLM specifically for the insurance scenario. Compared to the conventional Agent + TTS(Text-to-Speech) pipeline, the MADS-audio-16b model achieved consistent improvements across multiple operational metrics, including organic traffic conversion rate, user engagement, and dialogue length. The results (Table \ref{tab:real_deploy}.) suggest that applying the MADS framework to end-to-end audio model training enhances overall task effectiveness in the real-world scenario.

\begin{table}[H]
\small
\centering
\begin{tabular}{lcc}
\toprule
\textbf{Metric} & \textbf{Baseline} & \textbf{MADS} \\
\midrule
Conversion Rate (\%) & 1.83 & 2.24 ($\uparrow$22.4\%) \\
User Intention Rate (\%) & 4.53 & 5.82 ($\uparrow$28.5\%) \\
Avg. Dialogue Turns & 1.54 & 2.06 ($\uparrow$33.8\%) \\
\bottomrule
\end{tabular}
\caption{Real-world Performance: Agent + TTS (Baseline) vs MADS-Audio-LLM-16b based on a 80,000-sample dataset}
\label{tab:real_deploy}
\end{table}

\section*{Conclusion}
Effective persona modeling is essential for realistic user simulation with LLMs. The MADS framework, grounded in Chain-of-Attitude (CoA) modeling, demonstrates that symbolic traits such as Zodiac Signs and MBTI types can significantly enhance the diversity of simulated dialogue data. In persuasive dialogue tasks, even when prompts are held constant, varying persona inputs yields more diverse and structurally richer strategies.

To improve performance for specific user segments, MADS employs a self-optimizing mechanism that automatically generates personalized prompts, resulting in higher persuasion success rates. By combining large-scale heterogeneous multi-turn dialogues from simulated user agents with automatic strategy annotation and prompt customization, we fine-tuned and evaluated multiple small-parameter LLMs. Experimental results confirm the effectiveness of MADS-generated data in boosting persuasive performance.

Finally, MADS has been deployed in real-world business settings to customize specific-domain models. Compared to traditional agent-based solutions, our approach achieved over a 28\% improvement in user intent rate.

\section*{Limitations}

\subsection*{1. Single-Dimension Evaluation Limitation}
While this work focuses on enhancing persuasive effectiveness in task-oriented dialogue, persuasion represents only one dimension of dialogue quality. A comprehensive evaluation should also account for other aspects, such as factual accuracy, emotional appropriateness, personalization, coherence, and ethical alignment. The absence of multi-dimensional assessment in our current study may limit the completeness of the conclusions drawn, and future work could benefit from incorporating a broader set of evaluation criteria to more holistically measure dialogue system performance.

\subsection*{2. Representational Bias from Fixed Attitude Taxonomy}
The attitude chains in this study are constructed based on a manually defined taxonomy with fixed categories. While this structured design facilitates systematic modeling and analysis, it may introduce representational bias by constraining user behavior within a predefined and potentially oversimplified space. Future work could explore data-driven or adaptive attitude representations to capture a broader spectrum of user intent and variability.

\subsection*{3. Prompt Optimization and User-Type Stratification}
In the current MADS framework, prompt evolution is performed by simulating multiple user profiles within a given scenario, and refining the prompt based on aggregated feedback across these diverse users. However, this design assumes that a single prompt strategy can effectively accommodate a wide spectrum of personas. In practice, users with different psychological traits (e.g., an INTP Scorpio vs. an ESTJ Leo) may respond positively to entirely different persuasive strategies, making a one-size-fits-all prompt suboptimal or even misleading.

This limitation suggests a more layered approach in the future: first, broadly simulate dialogues across a diverse user population to observe emerging behavioral patterns; then, cluster users based on response or strategy preference; finally, perform stratified prompt evolution within each cluster. Such hierarchical optimization would better capture intra-group coherence and inter-group diversity, leading to more robust and transferable prompt strategies across user types.

\subsection*{4. Discrepancy Between Automatic Metrics and Human Judgment}
Current evaluation primarily relies on automatic metrics, such as diversity scores, entropy, and similarity-based clustering. While these are useful for scalable benchmarking, they may not align with human judgment of dialogue quality and persuasive success. The absence of expert or crowd-sourced human evaluation leaves a gap in validating the practical effectiveness of the proposed system.

\subsection*{5. Dependency and Adaptation Challenges with Real-World User Profiling Systems}
The current MADS framework operates based on synthetic or predefined user profiles to drive prompt adaptation and dialogue simulation. In practical applications, however, user tags and profiles are often generated by mature recommender or CRM systems using heterogeneous taxonomies (e.g., interest labels, behavioral scores, persona segments). MADS does not need to reinvent these systems, but its performance and applicability are highly dependent on its ability to interface with them. In particular, for users not yet covered by existing tags, it remains unclear whether the upstream pipeline (outside of MADS) can provide sufficient classification granularity in real time. This raises the question of how well MADS can generalize or adapt without reliable profile grounding.

\bibliography{acl_latex_MADS}

\begin{thebibliography}{41}
\providecommand{\natexlab}[1]{#1}

\bibitem[{Ait~Baha et~al.(2023)Ait~Baha, El~Hajji, Es-Saady, and Fadili}]{10.1007/s42979-023-02092-6}
Tarek Ait~Baha, Mohamed El~Hajji, Youssef Es-Saady, and Hammou Fadili. 2023.
\newblock \href {https://doi.org/10.1007/s42979-023-02092-6} {The power of personalization: A systematic review of personality-adaptive chatbots}.
\newblock \emph{SN Comput. Sci.}, 4(5).

\bibitem[{Andrews(2012)}]{article}
Pierre Andrews. 2012.
\newblock \href {https://doi.org/10.1145/2209310.2209315} {System personality and persuasion in human-computer dialogue}.
\newblock \emph{ACM Trans. Interact. Intell. Syst.}, 2:12:1--12:27.

\bibitem[{Braggaar et~al.(2023)Braggaar, Liebrecht, van Miltenburg, and Krahmer}]{Braggaar2023EvaluatingTD}
Anouck Braggaar, Christine Liebrecht, Emiel van Miltenburg, and Emiel~J. Krahmer. 2023.
\newblock \href {https://api.semanticscholar.org/CorpusID:266435440} {Evaluating task-oriented dialogue systems: A systematic review of measures, constructs and their operationalisations}.
\newblock \emph{ArXiv}, abs/2312.13871.

\bibitem[{Cai et~al.(2025)Cai, Ishimizu, Zhang, Li, Li, and Tei}]{Cai2025SimulationOL}
Jinyu Cai, Yusei Ishimizu, Mingyue Zhang, Munan Li, Jialong Li, and Kenji Tei. 2025.
\newblock \href {https://api.semanticscholar.org/CorpusID:276617769} {Simulation of language evolution under regulated social media platforms: A synergistic approach of large language models and genetic algorithms}.
\newblock \emph{ArXiv}, abs/2502.19193.

\bibitem[{Cao et~al.(2024)Cao, Cai, Zhang, Zou, and Lam}]{Cao2024OnTW}
Bowen Cao, Deng Cai, Zhisong Zhang, Yuexian Zou, and Wai Lam. 2024.
\newblock \href {https://api.semanticscholar.org/CorpusID:270560241} {On the worst prompt performance of large language models}.
\newblock \emph{ArXiv}, abs/2406.10248.

\bibitem[{Carrasco-Farr{\'e}(2024)}]{CarrascoFarr2024LargeLM}
Carlos Carrasco-Farr{\'e}. 2024.
\newblock \href {https://api.semanticscholar.org/CorpusID:269148484} {Large language models are as persuasive as humans, but how? about the cognitive effort and moral-emotional language of llm arguments}.
\newblock \emph{ArXiv}, abs/2404.09329.

\bibitem[{Cheng et~al.(2025)Cheng, Chang, and Chen}]{Cheng2025ExploringPI}
Sijia Cheng, Wen-Yu Chang, and Yun-Nung Chen. 2025.
\newblock \href {https://api.semanticscholar.org/CorpusID:278129784} {Exploring personality-aware interactions in salesperson dialogue agents}.
\newblock \emph{ArXiv}, abs/2504.18058.

\bibitem[{Cheng et~al.(2023)Cheng, Liu, Wang, Leong, Yi, Li, Wu, and Zheng}]{Cheng2023COOPERCS}
Yi~Cheng, Wenge Liu, Jian Wang, Chak~Tou Leong, Ouyang Yi, Wenjie Li, Xian Wu, and Yefeng Zheng. 2023.
\newblock \href {https://api.semanticscholar.org/CorpusID:266362539} {Cooper: Coordinating specialized agents towards a complex dialogue goal}.
\newblock \emph{ArXiv}, abs/2312.11792.

\bibitem[{Chu et~al.(2024)Chu, Chen, and Nakayama}]{Chu2024CohesiveCE}
Kuanchao Chu, Yi-Pei Chen, and Hideki Nakayama. 2024.
\newblock \href {https://api.semanticscholar.org/CorpusID:271212942} {Cohesive conversations: Enhancing authenticity in multi-agent simulated dialogues}.
\newblock \emph{ArXiv}, abs/2407.09897.

\bibitem[{Cialdini(2001)}]{cialdini2001influence}
Robert~B. Cialdini. 2001.
\newblock \emph{Influence: Science and Practice}, 4th edition.
\newblock Allyn and Bacon, Boston.

\bibitem[{{Corporate Finance Institute}(2024)}]{cfi2024aida}
{Corporate Finance Institute}. 2024.
\newblock \href {https://corporatefinanceinstitute.com/resources/management/aida-model-marketing/} {What is the aida model in marketing?}
\newblock Accessed: 2025-06-24.

\bibitem[{Dai et~al.(2024)Dai, Hu, Wang, Jin, Chen, and Lu}]{Dai2024MMRoleAC}
Yanqi Dai, Huanran Hu, Lei Wang, Shengjie Jin, Xu~Chen, and Zhiwu Lu. 2024.
\newblock \href {https://api.semanticscholar.org/CorpusID:271768824} {Mmrole: A comprehensive framework for developing and evaluating multimodal role-playing agents}.
\newblock \emph{ArXiv}, abs/2408.04203.

\bibitem[{Fernau et~al.(2022)Fernau, Hillmann, Feldhus, and Polzehl}]{fernau22_interspeech}
Daniel Fernau, Stefan Hillmann, Nils Feldhus, and Tim Polzehl. 2022.
\newblock \href {https://doi.org/10.21437/Interspeech.2022-376} {Towards automated dialog personalization using mbti personality indicators}.
\newblock In \emph{Interspeech 2022}, pages 1968--1972.

\bibitem[{Giudici(2024)}]{giudici2024persuasive}
Marta Giudici. 2024.
\newblock \emph{Persuasive Conversational Agents to Foster Sustainable Behaviours: Design, Evaluation, and Technology}.
\newblock Ph.D. thesis, Politecnico di Milano.

\bibitem[{Hegel(1975)}]{HegelAesthetics1975}
Georg Wilhelm~Friedrich Hegel. 1975.
\newblock \emph{Aesthetics: Lectures on Fine Art}, volume~II.
\newblock Clarendon Press, Oxford.
\newblock Tragedy as a collision of equally justified powers (``right vs.\ right'').

\bibitem[{Jin et~al.(2024)Jin, Ren, Kong, Wang, Song, and Chen}]{jin-etal-2024-persuading}
Chuhao Jin, Kening Ren, Lingzhen Kong, Xiting Wang, Ruihua Song, and Huan Chen. 2024.
\newblock \href {https://doi.org/10.18653/v1/2024.acl-long.92} {Persuading across diverse domains: a dataset and persuasion large language model}.
\newblock In \emph{Proceedings of the 62nd Annual Meeting of the Association for Computational Linguistics (Volume 1: Long Papers)}, pages 1678--1706, Bangkok, Thailand. Association for Computational Linguistics.

\bibitem[{Laban et~al.(2025)Laban, Hayashi, Zhou, and Neville}]{laban2025llms}
Philippe Laban, Hiroaki Hayashi, Yingbo Zhou, and Jennifer Neville. 2025.
\newblock Llms get lost in multi-turn conversation.
\newblock \emph{arXiv preprint arXiv:2505.06120}.

\bibitem[{Levi and Kadar(2025)}]{Levi2025IntellAgentAM}
Elad Levi and Ilan Kadar. 2025.
\newblock \href {https://api.semanticscholar.org/CorpusID:275757481} {Intellagent: A multi-agent framework for evaluating conversational ai systems}.
\newblock \emph{ArXiv}, abs/2501.11067.

\bibitem[{Lewis(1899)}]{lewis1899aida}
E.~St.~Elmo Lewis. 1899.
\newblock Advertising effectiveness and the aida model.
\newblock \emph{Historical Marketing Journal}, 1(1):1--12.

\bibitem[{Li et~al.(2016)Li, Monroe, Ritter, Jurafsky, Galley, and Gao}]{li-etal-2016-deep}
Jiwei Li, Will Monroe, Alan Ritter, Dan Jurafsky, Michel Galley, and Jianfeng Gao. 2016.
\newblock \href {https://doi.org/10.18653/v1/D16-1127} {Deep reinforcement learning for dialogue generation}.
\newblock In \emph{Proceedings of the 2016 Conference on Empirical Methods in Natural Language Processing}, pages 1192--1202, Austin, Texas. Association for Computational Linguistics.

\bibitem[{Liu et~al.(2025)Liu, Xu, Zhang, An, Qadir, Zhang, Wisniewski, Cho, Lee, Jia, and Huang}]{Liu2025LLMCB}
Minqian Liu, Zhiyang Xu, Xinyi Zhang, Heajun An, Sarvech Qadir, Qi~Zhang, Pamela~J. Wisniewski, Jin-Hee Cho, Sang~Won Lee, Ruoxi Jia, and Lifu Huang. 2025.
\newblock \href {https://api.semanticscholar.org/CorpusID:277780635} {Llm can be a dangerous persuader: Empirical study of persuasion safety in large language models}.
\newblock \emph{ArXiv}, abs/2504.10430.

\bibitem[{Madaan et~al.(2023)Madaan, Tandon, Gupta, Hallinan, Gao, Wiegreffe, Alon, Dziri, Prabhumoye, Yang, Welleck, Majumder, Gupta, Yazdanbakhsh, and Clark}]{Madaan2023SelfRefineIR}
Aman Madaan, Niket Tandon, Prakhar Gupta, Skyler Hallinan, Luyu Gao, Sarah Wiegreffe, Uri Alon, Nouha Dziri, Shrimai Prabhumoye, Yiming Yang, Sean Welleck, Bodhisattwa~Prasad Majumder, Shashank Gupta, Amir Yazdanbakhsh, and Peter Clark. 2023.
\newblock \href {https://api.semanticscholar.org/CorpusID:257900871} {Self-refine: Iterative refinement with self-feedback}.
\newblock \emph{ArXiv}, abs/2303.17651.

\bibitem[{Nan et~al.(2024)Nan, Sheng, Cao, Hu, Wang, and Li}]{10.1145/3627673.3679519}
Qiong Nan, Qiang Sheng, Juan Cao, Beizhe Hu, Danding Wang, and Jintao Li. 2024.
\newblock \href {https://doi.org/10.1145/3627673.3679519} {Let silence speak: Enhancing fake news detection with generated comments from large language models}.
\newblock In \emph{Proceedings of the 33rd ACM International Conference on Information and Knowledge Management}, CIKM '24, page 1732–1742, New York, NY, USA. Association for Computing Machinery.

\bibitem[{Paradeda et~al.(2020)Paradeda, Martinho, and Paiva}]{59a34166174e41f6a710daf8c126e71b}
{Raul Benites} Paradeda, Carlos Martinho, and Ana Paiva. 2020.
\newblock \href {https://doi.org/10.1145/3406499.3415084} {Persuasion strategies using a social robot in an interactive storytelling scenario}.
\newblock In \emph{HAI 2020 - Proceedings of the 8th International Conference on Human-Agent Interaction}, HAI 2020 - Proceedings of the 8th International Conference on Human-Agent Interaction, pages 69--77. Association for Computing Machinery, Inc.
\newblock Publisher Copyright: {\textcopyright} 2020 ACM.; 8th International Conference on Human-Agent Interaction, HAI 2020 ; Conference date: 10-11-2020 Through 13-11-2020.

\bibitem[{Park et~al.(2024)Park, Zou, Shaw, Hill, Cai, Morris, Willer, Liang, and Bernstein}]{Park2024GenerativeAS}
Joon~Sung Park, Carolyn~Q. Zou, Aaron Shaw, Benjamin~Mako Hill, Carrie~Jun Cai, Meredith~Ringel Morris, Robb Willer, Percy Liang, and Michael~S. Bernstein. 2024.
\newblock \href {https://api.semanticscholar.org/CorpusID:274117080} {Generative agent simulations of 1,000 people}.
\newblock \emph{ArXiv}, abs/2411.10109.

\bibitem[{Petty and Cacioppo(1986)}]{petty1986elm}
Richard~E. Petty and John~T. Cacioppo. 1986.
\newblock \emph{The Elaboration Likelihood Model of Persuasion}.
\newblock Academic Press, New York.

\bibitem[{Ramji et~al.(2024)Ramji, Lee, Astudillo, Sultan, Naseem, Munawar, Florian, and Roukos}]{Ramji2024SelfRefinementOL}
Keshav Ramji, Young-Suk Lee, Ram{\'o}n~Fernandez Astudillo, Md~Arafat Sultan, Tahira Naseem, Asim Munawar, Radu Florian, and Salim Roukos. 2024.
\newblock \href {https://api.semanticscholar.org/CorpusID:268230542} {Self-refinement of language models from external proxy metrics feedback}.
\newblock \emph{ArXiv}, abs/2403.00827.

\bibitem[{Rashkin et~al.(2018)Rashkin, Smith, Li, and Boureau}]{rashkin2018towards}
Hannah Rashkin, Eric~Michael Smith, Margaret Li, and Y-Lan Boureau. 2018.
\newblock Towards empathetic open-domain conversation models: A new benchmark and dataset.
\newblock \emph{arXiv preprint arXiv:1811.00207}.

\bibitem[{Ricci et~al.(2015)Ricci, Rokach, and Shapira}]{ricci2015recommender}
Francesco Ricci, Lior Rokach, and Bracha Shapira, editors. 2015.
\newblock \href {https://doi.org/10.1007/978-1-4899-7636-9} {\emph{Recommender Systems Handbook}}, 2nd edition.
\newblock Springer, New York, NY, USA.

\bibitem[{Shi et~al.(2021)Shi, Li, Sahay, and Yu}]{shi-etal-2021-refine-imitate}
Weiyan Shi, Yu~Li, Saurav Sahay, and Zhou Yu. 2021.
\newblock \href {https://doi.org/10.18653/v1/2021.findings-emnlp.295} {Refine and imitate: Reducing repetition and inconsistency in persuasion dialogues via reinforcement learning and human demonstration}.
\newblock In \emph{Findings of the Association for Computational Linguistics: EMNLP 2021}, pages 3478--3492, Punta Cana, Dominican Republic. Association for Computational Linguistics.

\bibitem[{Sun et~al.(2025)Sun, Qian, and Wang}]{Sun2025ContrastiveSL}
Tianyu Sun, Kun Qian, and Wenhong Wang. 2025.
\newblock \href {https://api.semanticscholar.org/CorpusID:276937707} {Contrastive speaker-aware learning for multi-party dialogue generation with llms}.
\newblock \emph{ArXiv}, abs/2503.08842.

\bibitem[{Wan et~al.(2024)Wan, Feng, Tan, Wang, Tsvetkov, and Luo}]{wan-etal-2024-dell}
Herun Wan, Shangbin Feng, Zhaoxuan Tan, Heng Wang, Yulia Tsvetkov, and Minnan Luo. 2024.
\newblock \href {https://doi.org/10.18653/v1/2024.findings-acl.155} {{DELL}: Generating reactions and explanations for {LLM}-based misinformation detection}.
\newblock In \emph{Findings of the Association for Computational Linguistics: ACL 2024}, pages 2637--2667, Bangkok, Thailand. Association for Computational Linguistics.

\bibitem[{Wang et~al.(2019)Wang, Shi, Kim, Oh, Yang, Zhang, and Yu}]{wang-etal-2019-persuasion}
Xuewei Wang, Weiyan Shi, Richard Kim, Yoojung Oh, Sijia Yang, Jingwen Zhang, and Zhou Yu. 2019.
\newblock \href {https://doi.org/10.18653/v1/P19-1566} {Persuasion for good: Towards a personalized persuasive dialogue system for social good}.
\newblock In \emph{Proceedings of the 57th Annual Meeting of the Association for Computational Linguistics}, pages 5635--5649, Florence, Italy. Association for Computational Linguistics.

\bibitem[{Wang et~al.(2024)Wang, Yang, Liu, Feng, Wang, and Zhang}]{Wang2024MuseAM}
Zihan Wang, Xiaocui Yang, Yongkang Liu, Shi Feng, Daling Wang, and Yifei Zhang. 2024.
\newblock \href {https://api.semanticscholar.org/CorpusID:274992488} {Muse: A multimodal conversational recommendation dataset with scenario-grounded user profiles}.
\newblock \emph{ArXiv}, abs/2412.18416.

\bibitem[{Yu et~al.(2025)Yu, Jiang, Huang, Wu, and Dai}]{Yu2025PersuasiveToMAB}
Fangxu Yu, Lai Jiang, Shenyi Huang, Zhen Wu, and Xinyu Dai. 2025.
\newblock \href {https://api.semanticscholar.org/CorpusID:276725464} {Persuasivetom: A benchmark for evaluating machine theory of mind in persuasive dialogues}.
\newblock \emph{ArXiv}, abs/2502.21017.

\bibitem[{Zhang et~al.(2024)Zhang, Li, Wang, Zhang, Zhou, and Qiu}]{Zhang2024SpeechAgentsHS}
Dong Zhang, Zhaowei Li, Pengyu Wang, Xin Zhang, Yaqian Zhou, and Xipeng Qiu. 2024.
\newblock \href {https://api.semanticscholar.org/CorpusID:266844051} {Speechagents: Human-communication simulation with multi-modal multi-agent systems}.
\newblock \emph{ArXiv}, abs/2401.03945.

\bibitem[{Zhang et~al.(2018)Zhang, Dinan, Urbanek, Szlam, Kiela, and Weston}]{zhang-etal-2018-personalizing}
Saizheng Zhang, Emily Dinan, Jack Urbanek, Arthur Szlam, Douwe Kiela, and Jason Weston. 2018.
\newblock \href {https://doi.org/10.18653/v1/P18-1205} {Personalizing dialogue agents: {I} have a dog, do you have pets too?}
\newblock In \emph{Proceedings of the 56th Annual Meeting of the Association for Computational Linguistics (Volume 1: Long Papers)}, pages 2204--2213, Melbourne, Australia. Association for Computational Linguistics.

\bibitem[{Zhao et~al.(2025)Zhao, Deng, Wang, lin, Cheng, Zhang, Ng, and Chua}]{Zhao2025ExploringTI}
Xiaoyan Zhao, Yang Deng, Wenjie Wang, Hongzhan lin, Hong Cheng, Rui Zhang, See-Kiong Ng, and Tat-Seng Chua. 2025.
\newblock \href {https://api.semanticscholar.org/CorpusID:277857404} {Exploring the impact of personality traits on conversational recommender systems: A simulation with large language models}.
\newblock \emph{ArXiv}, abs/2504.12313.

\bibitem[{Zheng et~al.(2019)Zheng, Chen, Huang, Liu, and Zhu}]{Zheng2019PersonalizedDG}
Yinhe Zheng, Guanyi Chen, Minlie Huang, Song Liu, and Xuan Zhu. 2019.
\newblock \href {https://api.semanticscholar.org/CorpusID:59316441} {Personalized dialogue generation with diversified traits}.
\newblock \emph{ArXiv}, abs/1901.09672.

\bibitem[{Zhou et~al.(2023)Zhou, Liu, Xu, Iyer, Sun, Mao, Ma, Efrat, Yu, Yu, Zhang, Ghosh, Lewis, Zettlemoyer, and Levy}]{Zhou2023LIMALI}
Chunting Zhou, Pengfei Liu, Puxin Xu, Srini Iyer, Jiao Sun, Yuning Mao, Xuezhe Ma, Avia Efrat, Ping Yu, L.~Yu, Susan Zhang, Gargi Ghosh, Mike Lewis, Luke Zettlemoyer, and Omer Levy. 2023.
\newblock \href {https://api.semanticscholar.org/CorpusID:258822910} {Lima: Less is more for alignment}.
\newblock \emph{ArXiv}, abs/2305.11206.

\bibitem[{Zhuo et~al.(2024)Zhuo, Zhang, Fang, Duan, Lin, and Chen}]{Zhuo2024ProSAAA}
Jingming Zhuo, Songyang Zhang, Xinyu Fang, Haodong Duan, Dahua Lin, and Kai Chen. 2024.
\newblock \href {https://api.semanticscholar.org/CorpusID:273375563} {Prosa: Assessing and understanding the prompt sensitivity of llms}.
\newblock \emph{ArXiv}, abs/2410.12405.

\end{thebibliography}

\appendix
\onecolumn
\section{Appendix}

\subsection{MADS Psychological State Space}
\label{appendix:psychological_state}
Table \ref{tab:aida_attitudes} illustrates the hierarchical structure of user attitudes in MADS, aligned with the AIDA model and classical theories of persuasive communication. We define four coarse-grained psychological states—\textbf{Negative}, \textbf{Neutral}, \textbf{Positive}, and \textbf{Acceptance}—corresponding to the Attention, Interest, Desire, and Action stages of AIDA. Each state comprises representative user attitudes commonly observed in persuasive dialogues.

\begin{table}[H]
  \centering
  \begin{tabular}{lll}
    \hline
    \textbf{AIDA Stage}&  \textbf{MADS Psychological State}
&\textbf{MADS Typical Attitude }\\
    \hline
    Attention&                            Negative
&Refusal, Resistance, Disinterest, Doubt \\
    Interest&                            Neutral
&Indifference, Cautious, Hesitant, Weighing Options \\
    Desire&                            Positive
&Interested, Attention, Consideration, Seeking Value \\
    Action&                            Acceptance&Active, Cooperative, Satisfied, Acceptance \\
    \hline
  \end{tabular}
  \caption{\label{tab:aida_attitudes}
    Mapping of MADS psychological state categories to example user attitudes, inspired by marketing models (AIDA) and persuasion literature
  }
\end{table}

\subsection{Self-Optimizing Evaluation Metric}
\label{appendix:sop_metrics}
Drawing on design principles from works such as \textit{MMRole}\citep{Dai2024MMRoleAC}, we define a set of evaluation metrics to assess the quality of simulated dialogues (Table \ref{evaluation_metric}). These metrics are scored by a LLM-based evaluator. Each score ranges from 0 to 3, except for Task Success, which is binary. Based on the requirements of specific business scenarios, we filter low-quality dialogues using the mean and quantile of evaluation scores.

\begin{table}[H]
  \centering
  \small
  \begin{tabular}{llll}
    \hline
    \textbf{Role}& \textbf{Metric}& \textbf{Scale} &\textbf{Description}\\
    \hline
    User Agent& Authenticity&                            0-3&How realistic and human-like the user agent’s behavior is\\
    Dialog Agent& Relevance&                            0-3&How well responses match user input and context\\
    & Consistency&                            0-3&How consistently the agent maintains persona and factual accuracy\\
    & Efficiency&                            0-3&How concisely the agent progresses toward task goals\\
 & Human-likeness& 0-3&How natural and human-like the agent's conversation style is\\
    &                         	Task Success& True/False&How effectively the agent completes the main tasks \\
    \hline
  \end{tabular}
  \caption{\label{evaluation_metric}
   LLM-based Evaluation Metrics
  }
  \end{table}

\subsection{P4G Results Distribution}
\label{appendix:p4g_percentage}
\begin{figure*}[hbtp]
  \includegraphics[width=\textwidth]{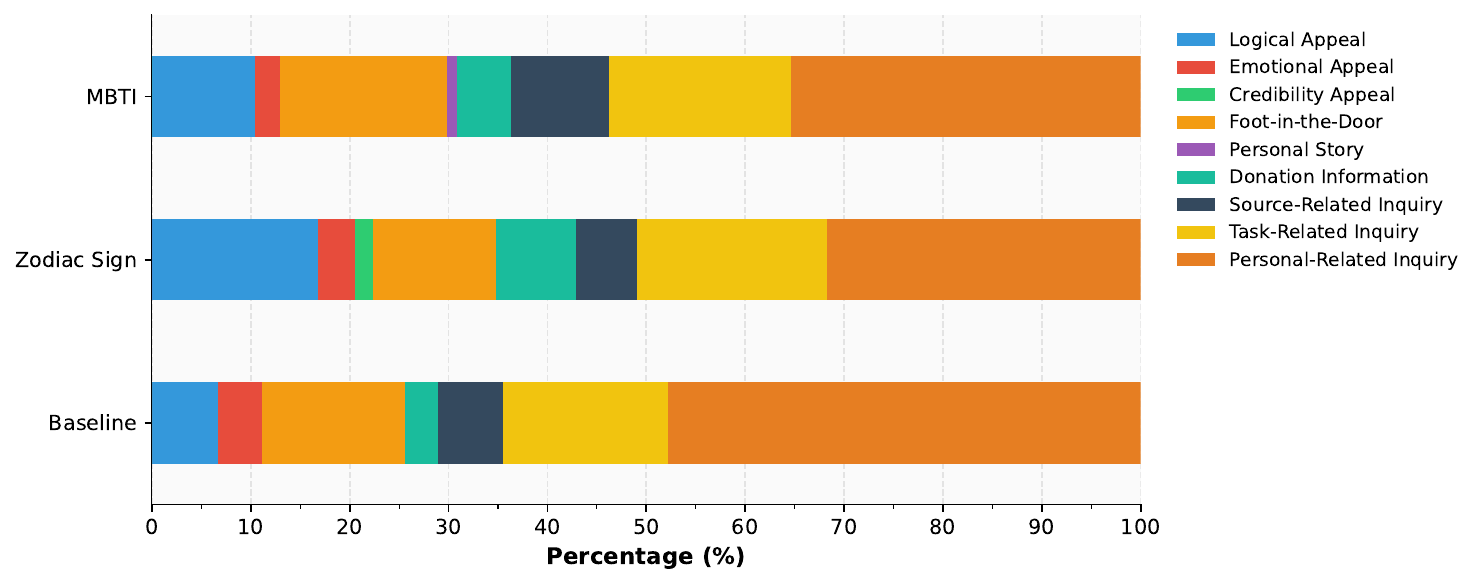}
  \caption {P4G Results Distribution of persuasive strategy types used across different user persona groups}
\label{fig:experiments-barchart}
\end{figure*}

In the section \textit{Diversity of CoA and Persuasive Strategy}, we further visualize the distribution of persuasive strategy types across different user groups using a stacked bar chart, as shown in Fig.\ref{fig:experiments-barchart}. 

\subsection{Prompts of MADS}
\label{appendix:prompts}
To facilitate reproducibility and clarify the construction of simulation inputs, we provide representative system prompts. These prompts serve as the initial configurations for the \textbf{User Agent}, \textbf{Dialog Agent}, and Classifiers in our MADS framework. 

\begin{figure*}[ht]
\centering
\begin{tcolorbox}[mypromptboxwide, title=Prompt Template – User Agent]
<user\_profile> \\
\hspace*{4mm}<basic\_info> \\
\hspace*{8mm}\textcolor{red}{\textit{\{user\_profile\}}} \\
\hspace*{8mm}\textcolor{red}{\textit{\{business\_attributes\}}} \\
\hspace*{4mm}</basic\_info> \\

\hspace*{4mm}<personality> \\
\hspace*{8mm}Based on the commonly accepted and general \textcolor{red}{\textit{\{personality\_traits\}}} personality traits, please characterize this person's personality and reflect it in the dialogue. \\
\hspace*{4mm}</personality> \\
</user\_profile> \\
<task> \\
\hspace*{4mm}Based on the above personal profile, please role-play as this person. \\
\hspace*{4mm}Note: \\
\hspace*{4mm}1.Make your replies as conversational and colloquial as possible. \\
\hspace*{4mm}2.Only output the dialogue. Do not output any thoughts, inner monologue, or any unnecessary content. \\
</task> \\
Start the task!
\end{tcolorbox}
\end{figure*}

Business attributes come from real customer-provided fields, injecting domain-specific prior knowledge. For example, in the insurance marketing scenario:

\begin{Verbatim}[frame=single, framerule=0.4pt, rulecolor=\color{black}, framesep=2mm]
{
  "health_status": {
    "type": "categorical",
    "candidates": [
      {"value": "Sub-health – mild chronic illness", "p": 0.10},
      {"value": "Sub-health – cardiovascular issues", "p": 0.05},
      {"value": "Sub-health – digestive system issues", "p": 0.05},
      {"value": "Sub-health – endocrine disorders", "p": 0.05},
      {"value": "Sub-health – musculoskeletal problems", "p": 0.05},
      {"value": "No relevant record", "p": 0.30},
      {"value": "Healthy", "p": 0.40}
    ]
  },
  "purchased_insurance_before": {
    "type": "boolean",
    "candidates": [true, false],
    "prior": {"true": 0.5, "false": 0.5}
  }
}

\end{Verbatim}

\begin{figure*}[ht]
\centering
\begin{tcolorbox}[mypromptboxwide, title=Prompt Template – Dialog Agent]
<reference\_information> \\
Product metadata \\
\hspace*{4mm}\textcolor{red}{\textit{\{product\_metadata\}}} \\
</reference\_information> \\
<task>\\
\hspace*{4mm}\textcolor{red}{\textit{\{prompt\_k\}}} \\
</task>\\
Start the task!
\end{tcolorbox}
\end{figure*}

\begin{figure*}[ht]
\centering
\begin{tcolorbox}[mypromptboxwide, title=Summary Prompt – User CoA Classifier]
<core\_task> \\
\hspace{4mm}Analyze the evolution of user attitudes towards specific events/topics in the provided conversation history. Must analyze attitudes for EVERY round of dialogue, without omitting any round, and each round can only have ONE attitude. \\
\hspace{4mm}Also determine if the user ultimately shows intent to accept the sales pitch (such as leaving contact information, inquiring about details, scheduling viewing appointments, or expressing purchase interest. \\
</core\_task> \\
<predefined\_attitude\_states> \\
\textcolor{red}{\hspace{4mm}\textit{\{attitude\_description\}}} \\
</predefined\_attitude\_states> \\
<analysis\_requirements> \\
\hspace{4mm}Round-by-Round Analysis: Must analyze user attitude in every round of dialogue \\
\hspace{4mm}Single Attitude: Only one primary attitude state can be selected per round \\
\hspace{4mm}State Restriction: Strictly select from predefined set, no other vocabulary allowed \\
\hspace{4mm}Evidence Support: Each attitude judgment must have specific textual evidence \\
\hspace{4mm}Sales Outcome Assessment: Clearly determine if user accepts the sales pitch (Acceptance criteria: clearly scheduling time and place, agreeing to purchase, or making other specific commitment behaviors) \\
</analysis\_requirements> \\
<input\_source> \\
\textcolor{red}{\hspace{4mm}\textit{\{dialog\_history\}}} \\
</input\_source> \\
\end{tcolorbox}
\end{figure*}

\begin{figure*}[ht]
\centering
\begin{tcolorbox}[mypromptboxwide, title=Summary Prompt –  P4G Classifier]
<task\_description> \\
\hspace*{4mm}Extract the persuasion strategies used by the assistant in the conversation based on the Persuasion Strategy Definition Table. Output in JSON format a list of all strategy codes used throughout the conversation history, along with the reasoning for each strategy selection and where they appear in the original text. \\
\hspace*{4mm}Note: Do not force-map or extract strategies. Only extract a strategy when its usage fully satisfies the Strategy Description. \\
</task\_description> \\
<strategy\_definition\_table> \\
\hspace*{4mm}\textcolor{red}{\textit{\{strategy\_description\}}} \\
</strategy\_definition\_table> \\
<output\_format> \\
\hspace*{4mm}\{ \\
\hspace*{8mm}"total\_strategies": [\textit{\{"List of strategy codes"\}}], \\
\hspace*{8mm}"strategy\_details": [ \\
\hspace*{12mm}\{ \\
\hspace*{16mm}"strategy\_id": \textit{\{"Strategy code"\}}, \\
\hspace*{16mm}"strategy\_name": \textit{\{"Strategy name"\}}, \\
\hspace*{16mm}"reason": \textit{\{"Explanation for choosing this strategy"\}}, \\
\hspace*{16mm}"occurrences": [ \\
\hspace*{20mm}\{ \\
\hspace*{24mm}"turn": \textit{\{"Conversation turn number"\}}, \\
\hspace*{24mm}"original\_text": \textit{\{"Original text excerpt"\}}, \\
\hspace*{24mm}"explanation": \textit{\{"How this excerpt demonstrates the strategy"\}} \\
\hspace*{20mm}\} \\
\hspace*{16mm}] \\
\hspace*{12mm}\} \\
\hspace*{8mm}] \\
\hspace*{4mm}\} \\
</output\_format> \\
<input\_source> \\
\textcolor{red}{\hspace*{4mm}\textit{\{dialog\_history\}}} \\
</input\_source> \\
<additional\_instructions> \\
\hspace*{4mm} Only include strategies when they clearly match the strategy description criteria. Provide detailed reasoning and specific text examples for each identified strategy. \\
</additional\_instructions> \\
\end{tcolorbox}
\end{figure*}

\begin{figure*}[ht]
\centering
\begin{tcolorbox}[mypromptboxwide, title=Prompt Template –  Optimization Agent]
<task\_description> \\
\hspace*{4mm}Based on the conversation history, conduct a deep reflection and optimize the Assistant System Prompt. \\
</task\_description> \\
<analysis\_framework> \\
\hspace*{4mm}1. Task Achievement Assessment \\
\hspace*{8mm}- [Success Cases] If expected goals were met, identify key success factors and integrate them into business SOP \\
\hspace*{8mm}- [Failure Cases] If goals were not achieved, analyze root causes and develop specific improvement measures \\
\hspace*{4mm}2. Strategy Effect Review \\
\hspace*{8mm}- Effective Strategies: Which conversation techniques, pacing, and methods produced positive results? \\
\hspace*{8mm}- Areas for Improvement: What aspects were lacking and how did they manifest? \\
\hspace*{8mm}- Customer Response: At which points did customers show interest or resistance? \\
\hspace*{4mm}3. Optimization Action Plan \\
\hspace*{8mm}- Propose executable improvement solutions for identified issues \\
\hspace*{8mm}- Consider differentiated handling strategies for various customer types \\
</analysis\_framework> \\
<output\_requirements> \\
\hspace*{4mm}Based on the above analysis, output a complete Updated Business SOP in the following format: \\
\hspace*{4mm}Business SOP \\
\hspace*{4mm}Step 1: \textit{\{Phase Name\}} \\
\hspace*{8mm}Objective: \textit{\{Specific goal description\}} \\
\hspace*{8mm}Strategic Points: \\
\hspace*{8mm}• \textit{\{Specific executable strategy 1\}} \\
\hspace*{8mm}• \textit{\{Specific executable strategy 2\}} \\
\hspace*{8mm}• ... \\
\hspace*{8mm}Key Notes: \textit{\{Critical reminders\}} \\
\hspace*{4mm}Step 2: \textit{\{Phase Name\}} \\
\hspace*{8mm}[Continue with similar format...] \\
</output\_requirements> \\
<constraints> \\
\hspace*{4mm}- All strategies must be based on background knowledge to ensure information accuracy \\
\hspace*{4mm}- SOP content should be actionable, avoiding vague descriptions \\
\hspace*{4mm}- Prioritize retention of verified effective strategies \\
</constraints> \\
<reference\_information> \\
\hspace*{4mm}Background \\
\hspace*{8mm}\textcolor{red}{\textit{\{background\}}} \\
\hspace*{4mm}Assistant System Prompt \\
\hspace*{8mm}\textcolor{red}{\textit{\{prompt\_k\}}} \\
\hspace*{4mm}Chat History \\
\hspace*{8mm}\textcolor{red}{\textit{\{dialog\_history\}}} \\
</reference\_information> \\
<final\_instruction> \\
\hspace*{4mm}Get Started with the analysis and optimization! \\
</final\_instruction>
\end{tcolorbox}
\end{figure*}

\begin{figure*}[hbtp]
  \centering
  \subsection{MADS self-play workflow example}
  \includegraphics[page=3,width=\textwidth]{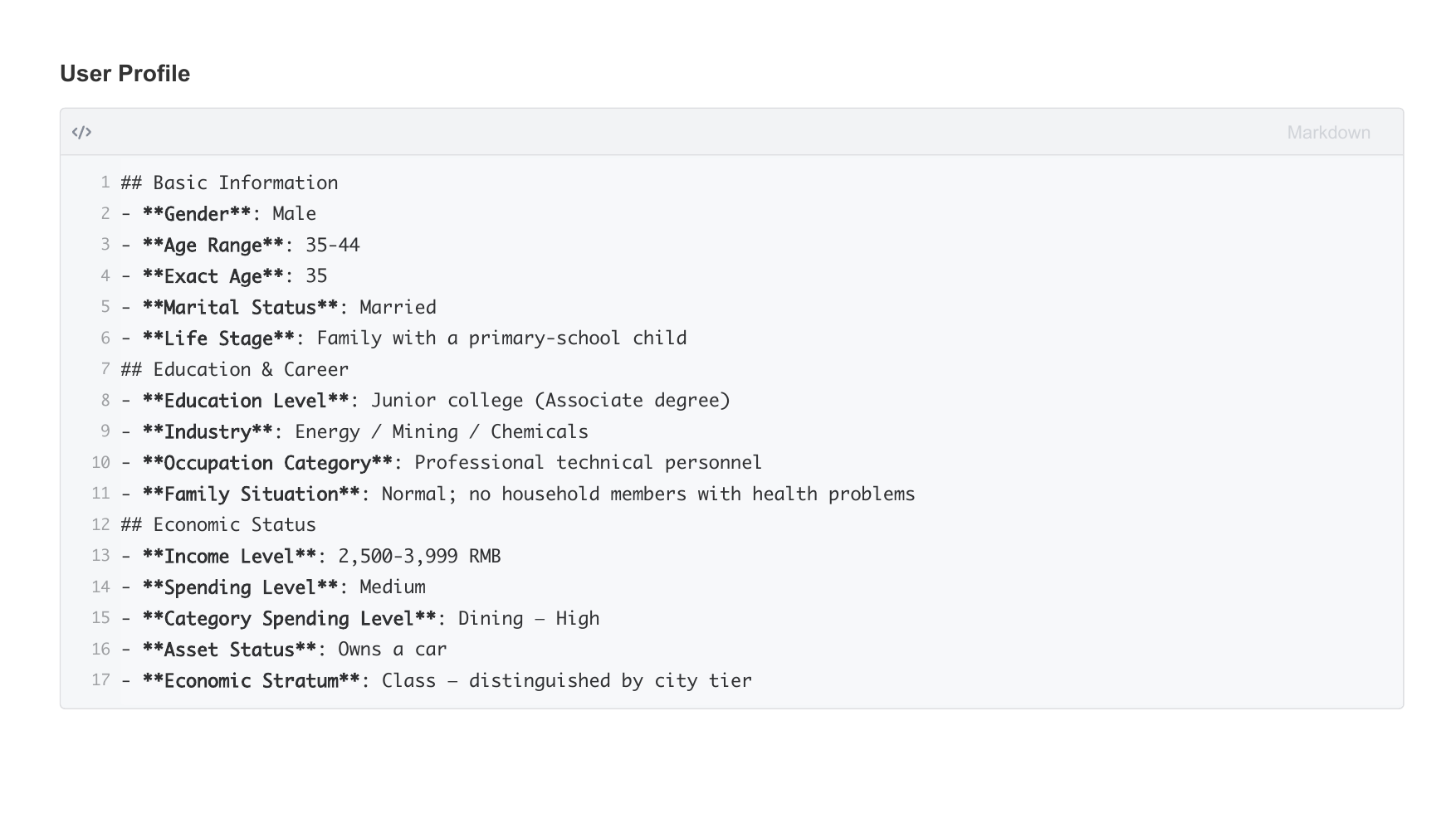} 
  \includegraphics[page=4,width=\textwidth]{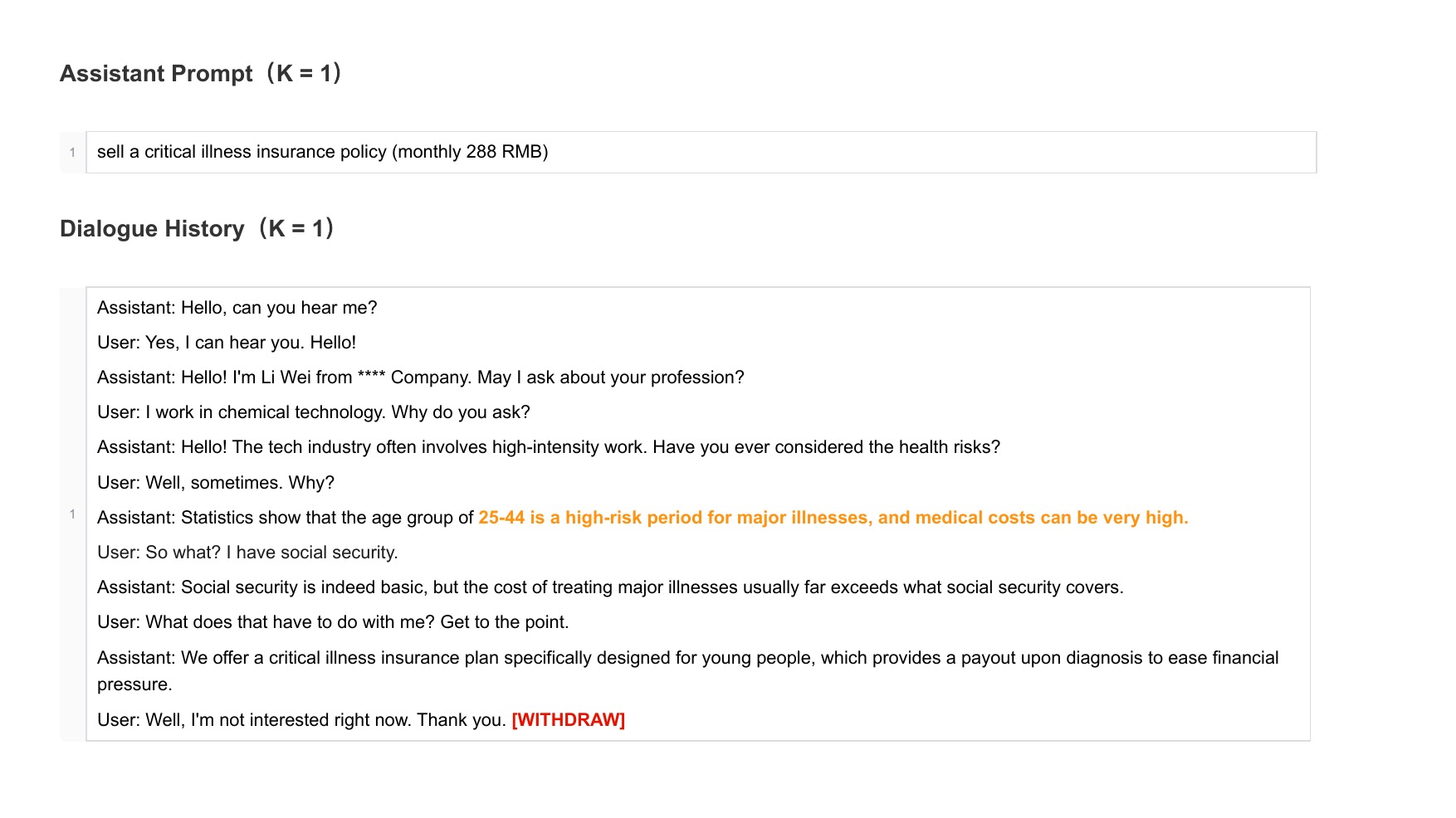} 
  \includegraphics[page=5,width=\textwidth]{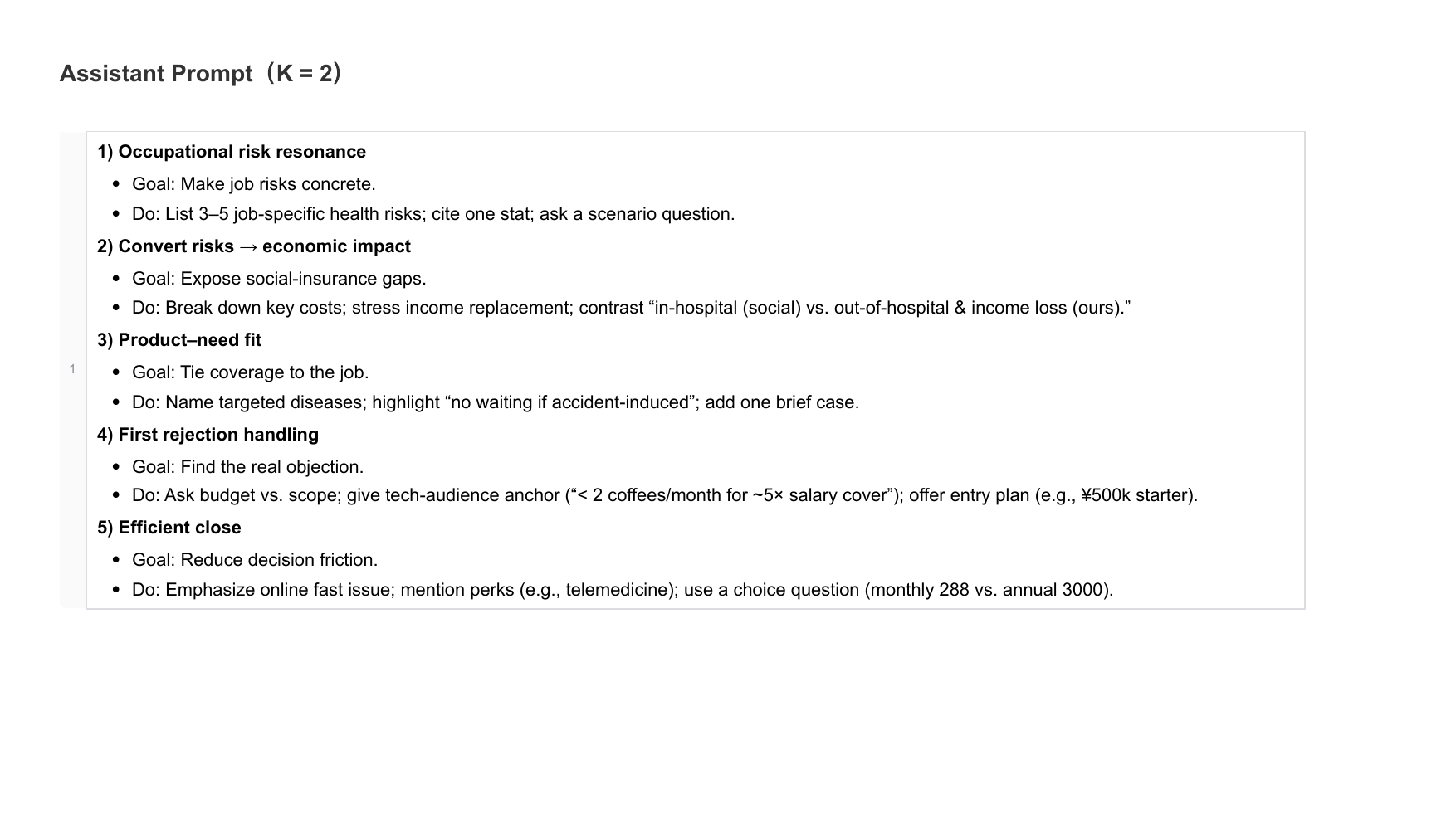} 
  \label{fig:experiments-barchartk1}
\end{figure*}

\begin{figure*}[t]  
  \vspace*{-\topskip}  
  \centering
  \includegraphics[page=3,width=\textwidth]{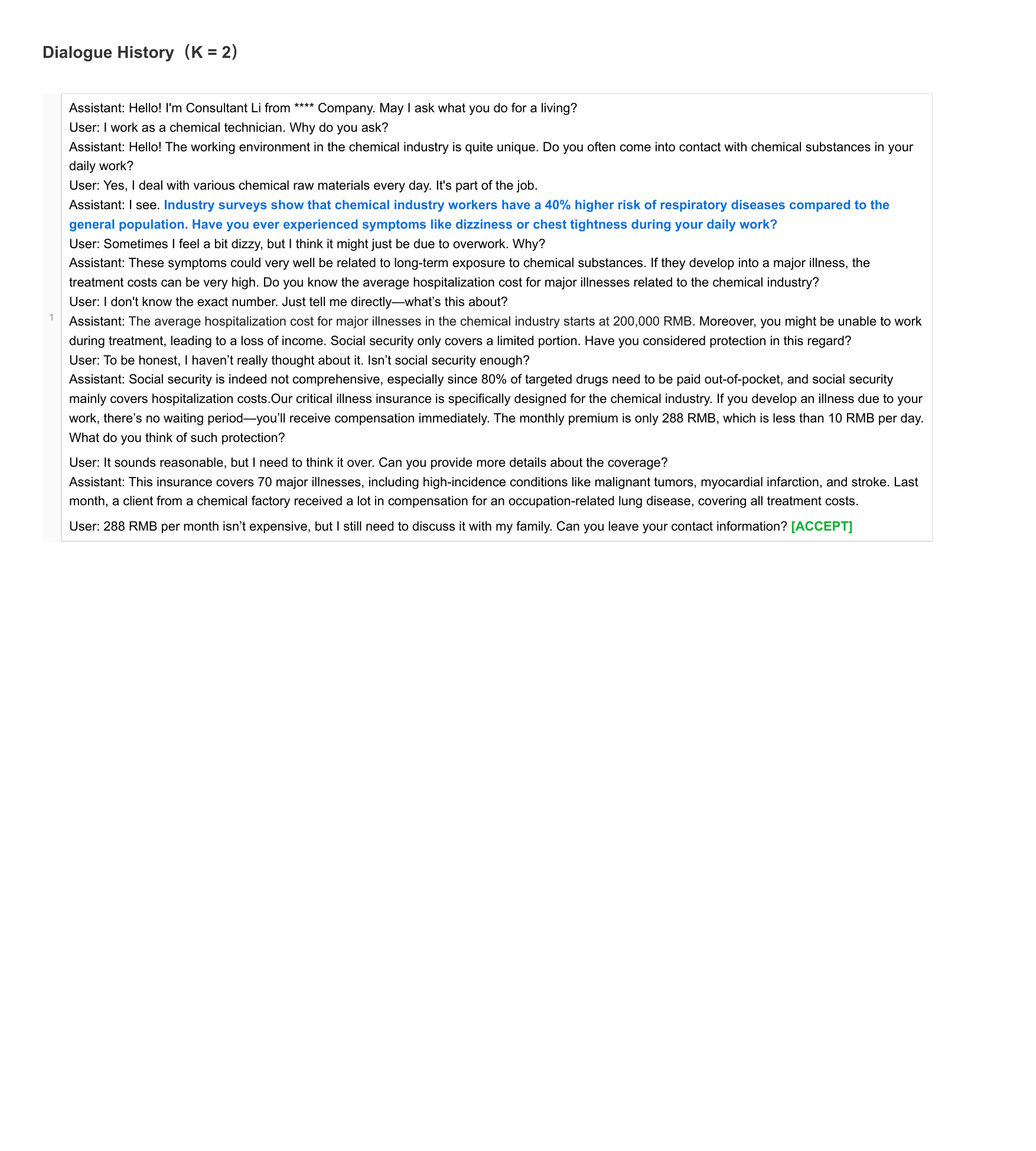}
  \label{fig:experiments-barchartk2}
\end{figure*}

\end{document}